\documentclass[sigconf]{acmart}

\usepackage{graphicx} 
\usepackage{hyperref}

\usepackage{graphicx} 
\usepackage{caption} 

\newcommand{\Pm}{{P}}      

\usepackage{tikz}
\usetikzlibrary{arrows.meta, positioning}

\def\ifundefined#1{\expandafter\ifx\csname#1\endcsname\relax}

\long\def\comment#1{}
\newcommand{\Df}[1]{{\ifvmode\else\/\fi \em #1\/\index{#1}}}

\usepackage{subcaption} 
\usepackage{amsmath}

\newcommand{\MerxOff}[1]{}

\usepackage{listings}
\AtBeginDocument{%
  }

\setcopyright{acmlicensed}
\begin{document}

\title{Confounding is a Pervasive Problem in Real World Recommender Systems}

\author{Alexander Merkov}

\author{David Rohde}
\author{Alexandre Gilotte}
\author{Benjamin Heymann}

\affiliation{%
  \institution{Criteo}
  \city{Paris}
  \country{France}
}

\begin{abstract}
    Unobserved confounding arises when an unmeasured feature influences both the treatment and the outcome, leading to biased causal effect estimates. This issue undermines observational studies in fields like economics, medicine, ecology or epidemiology. Recommender systems leveraging fully observed data  seem not to be vulnerable to this problem.  However many standard practices in recommender systems result in observed features being ignored, resulting in effectively the same problem.
    This paper will show that numerous common practices such as feature engineering, A/B testing and modularization can in fact introduce confounding into  recommendation systems and hamper their performance.   Several illustrations of the phenomena are provided, supported by simulation studies with practical suggestions about how practitioners may reduce or avoid the affects of confounding in real systems.
\end{abstract}

\maketitle

\section{Introduction}

In a recommender system all information that leads to a recommendation is known, which in principle should make unobserved confounding impossible.  However this paper shows that many common practices for training click models in production results in available  covariates or features being ignored.  These practices are pervasive in real systems and likely result in many systems behaving sub-optimally.

\begin{figure}[h]
    \centering
    \includegraphics[width=0.4\textwidth]{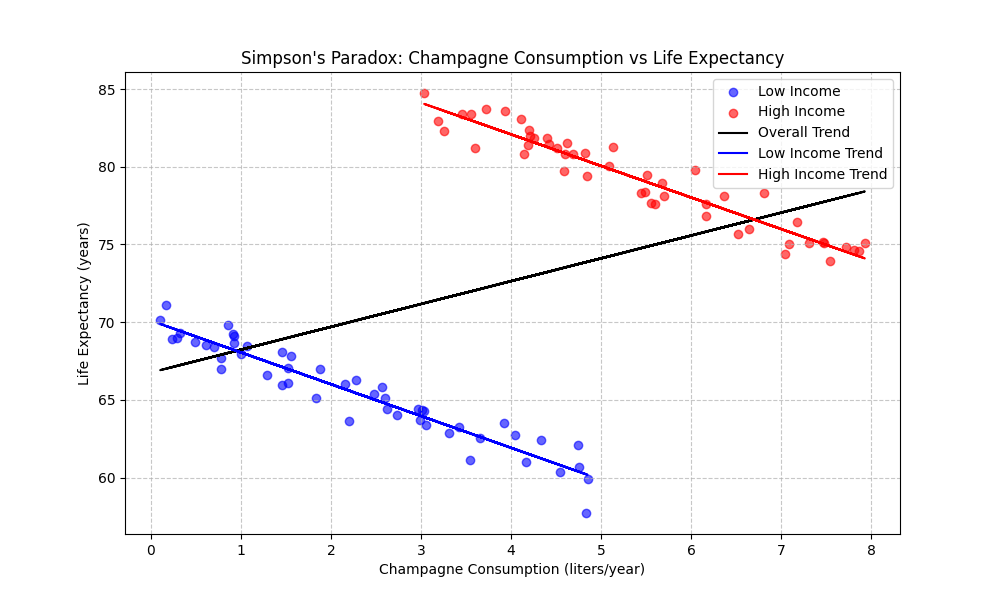} 
    \caption{Simpson's Paradox: Life expectancy increases with Champagne consumption overall, but decreases with Champagne consumption for a fixed income.}
    \label{fig:champagne}
\end{figure}

Researchers in recommender systems are probably somewhat familiar with the way confounding can lead to scenarios where correlation does not equal causation, but to set concepts and terminology Figure \ref{fig:champagne} provides a simple illustration of  Simpson's famous paradox.  The diagram serves to illustrate how three variables interact, life expectancy, income and Champagne consumption.  The diagram shows that life expectancy increases with Champagne consumption, but it also illustrates that once income is adjusted for,  life expectancy decreases with Champagne consumption  (which is perhaps a more plausible causal relationship).  Simpson's paradox refers to the fact that there can be different causal interpretations between three or more variables as in this example.  A confounder is a covariate that must be incorporated into the model in order for the correct causal inference to be arrived at (in this case income is a confounder), in natural experiments it is possible (indeed common) for the confounder to be unobserved rendering a correct causal inference impossible.   If a click model is used to produce personalized recommendations then in principle everything is observed and confounding will not occur.  
Many `best practices' in production systems can result in observed information being ignored resulting in confounding and reduced performance.

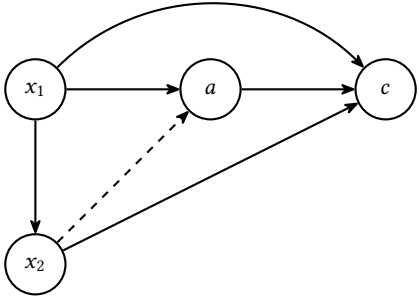
\begin{figure}[h]
\centering
\begin{tikzpicture}
    [node distance=1.2cm and 1.5cm, auto, >={Stealth[round]}, thick,
     every node/.style={circle, draw, fill=white, minimum size=8mm, inner sep=2pt}]
    \node(x1) {$x_1$};
    \node[right=of x1] (a) {$a$};
    \node[right=of a] (c) {$c$};
    \node[below=1.5cm of x1] (x2) {$x_2$};
    \draw[->] (x1) -- (a);
    \draw[->] (x1) to[out=45,in=135] (c);
    \draw[->] (x2) -- (c);
    \draw[->] (x1) -- (x2);    
    \draw[->, dashed] (x2) to[out=45,in=225] (a);
    \draw[->] (a) -- (c);
\end{tikzpicture}
\caption{Causal DAG: Solid edges represent the base graph; the dashed edge \( x_2 \to a \) is optional.}
\label{fig:graph}
\end{figure}

Let a click be denoted $c$, the recommendation or action $a$, and the covariate used for personalization currently is $x_1$,  a second covariate that may also be used to personalize is denoted $x_2$,   The terms covariate and feature are used interchangeably.  The causal graph is shown in Figure \ref{fig:graph}, the click $c$ is caused by all three of the action $a$ and the covariates $x_1$ and $x_2$.  In the current setup the recommendation is personalized only by $x_1$ only indicated by the arrow from $x_1 \rightarrow a$, the dashed arrow from $x_2 \rightarrow a$ is absent (for now).  The joint distribution on the covariates is denoted by the arrow $x_1 
\rightarrow x_2$, this choice of direction simplifies the proofs.  The recommender system is trained on a log of observations of $a,x_1,x_2,c$ by regressing $c$ on $x_1,a$ i.e. a logistic regression model is fit:

\begin{align}
 \hat{\beta} =   ~{\rm argmax}_\beta \sum_{i=1}^n c_i & \log \sigma((x_{1_i} \otimes a_i)^T \beta) \\ 
&  +  (1-c_i) \log (1-\sigma(( x_{1_i} \otimes a_i)^T \beta)) \nonumber
\label{ignorex2}    
\end{align}

\noindent
where $\sigma(\cdot)$ is the logistic sigmoid and $\otimes$ is the Kronecker product.  It is assumed that training on recent history (the previous day) mitigates any non-stationarity.  This model can then be used in order to produce a new epsilon greedy recommendation policy, where the exploration parameter is given by $\epsilon$ and the number of actions is $A$.

\[
\pi(a|x_1) = (1-\epsilon) \boldsymbol{1} \{a ={\rm argmax_{a'} } (x_{1} \otimes a')^T \hat{\beta}  \} + \frac{\epsilon}{A}.
\]

\noindent
We verify that this standard practice is in fact sound from a causal inference point of view.   Applying the do calculus gives the following expression:

\begin{align}
P(c|{\rm do}(a), x_1) = \sum_{x_2} P(c|a, x_1,x_2)P(x_2|x_1) 
\label{docalc}    
\end{align}

\noindent
however the absent of the edge  $x_2 \rightarrow a$, by the backdoor criterion shows that $x_2$ is ignorable.  This means that the logistic regression model in Equation 1, directly estimates the causal quantity in Equation 2.




What if the system then changes such that personalization now uses $x_2$ i.e. the edge $x_2 
\rightarrow a$ is added, but the goal remains to propose a new system that \emph{does not use} $x_2$.  Applying the do calculus still gives the same formula for the causal effect (Equation \ref{docalc}), however now ignorability of $x_2$ no longer applies and this formula must be applied (which is rather unpalatable to practitioners as it involves fitting a larger model to both $P(c|a,x_1,x2)$ and $P(x_2|x_1)$ and performing a sum or integral over $x_2$).

In a recommender system  $x_1$ is the feature currently used for personalization and $x_2$ is an additional feature that may be used in the future.  This paper makes two main points:

\begin{itemize}
    \item If the recommender system ignores some feature $x_2$ i.e. there is no arrow $x_2 \rightarrow a$, then ignorability applies even if $x_2 \rightarrow c$.  Researchers do not need to worry about confounding or causal theory in this case.
    \item Numerous common practices result in the link between some feature $x_2 \rightarrow a$ existing, but the model training procedure ignoring this link and erroneously assume $x_2$ to be ignorable.  
    This mistake is easy to make, because a methodology that was sound when there was no edge $x_2\rightarrow a$ becomes broken.
\end{itemize}

\noindent
The main mathematical results that this paper relies on is that Equation \ref{docalc} is the appropriate formula for both causal graphs in Figure 
\ref{fig:graph} (both with and without the dashed line), as can be shown using the do calculus\footnote{By  application of the do calculus' Rule 2 and Rule 3, resulting in a rule that slightly generalized the backdoor rule \cite{pearl1995causal}.}.  Secondly that ignorability of $x_2$ applies if and only if the graph in Figure 1 has no arrow $x_2 \rightarrow a$, as can be shown by applying the backdoor criterion or the Rubin Causal Model\footnote{When the arrow \( x_2 \to a \) is present in the causal graph (\( x_1 \to a, x_2 \to a, x_1 \to c, x_2 \to c, a \to c, x_1 \to x_2 \)), \( x_2 \) is not ignorable for estimating \( P(c \mid x_1, \mathrm{do}(a)) \), as it confounds the \( a \to c \) relationship via the backdoor path \( a \leftarrow x_2 \to c \), requiring adjustment with \( \sum_{x_2} P(c \mid x_1, x_2, a) P(x_2 \mid x_1) \). When the arrow \( x_2 \to a \) is absent (\( x_1 \to a, x_1 \to c, x_2 \to c, a \to c, x_1 \to x_2 \)), \( x_2 \) is ignorable, as all backdoor paths (\( a \leftarrow x_1 \to c \), \( a \leftarrow x_1 \to x_2 \to c \)) are blocked by conditioning on \( x_1 \), yielding \( P(c \mid x_1, \mathrm{do}(a)) = P(c \mid x_1, a) \).  Similarly, under the Rubin Causal model when the arrow \( x_2 \to a \) is present in the causal graph (\( x_1 \to a, x_2 \to a, x_1 \to c, x_2 \to c, a \to c, x_1 \to x_2 \)), \( x_2 \) is not ignorable for estimating \( P(c \mid x_1, \mathrm{do}(a)) \), as it confounds the \( a \to c \) relationship via the backdoor path \( a \leftarrow x_2 \to c \), requiring adjustment with \( \sum_{x_2} P(c \mid x_1, x_2, a) P(x_2 \mid x_1) \). When the arrow \( x_2 \to a \) is absent (\( x_1 \to a, x_1 \to c, x_2 \to c, a \to c, x_1 \to x_2 \)), \( x_2 \) is ignorable, as all backdoor paths (\( a \leftarrow x_1 \to c \), \( a \leftarrow x_1 \to x_2 \to c \)) are blocked by conditioning on \( x_1 \), yielding \( P(c \mid x_1, \mathrm{do}(a)) = P(c \mid x_1, a) \), for ignorability in the Rubin Causal Model see \cite{rubin1974estimating,rosenbaum1983central}.
}.
There are two main cases where a link might be introduced into $x_2\rightarrow a$, yet is ignored in subsequent training.  The first is due to feature engineering, the second is due to modularization (both good practices from other points of view).



\section{Confounding due to Feature Engineering and A/B Testing}

Feature engineering is a standard optimization of machine learning models in recommendation teams.  Consider the following story.

\paragraph{Day 0:} the recommender system is doing pure random exploration of the action $a$ independent of all features.  That is the policy is $\pi_0(a|x_1,x_2)=\pi_0(a)=\frac{1}{A}$.  Data is collected on Day 0 i.e. $\mathcal{D}_0 = \{c^{(i)},x_1^{(i)},x_2^{(i)},a^{(i)})_{i=1}^{L_0}$.  

The reco team decides that features $x_1$ are more interesting, and they will ignore $x_2$ for now, they fit the model with maximum likelihood i.e.

\[
\hat{\beta}_0 = {\rm argmax}_\beta \sum_{i=1}^{L_0} \log \Pm(c^{(i)}|x_1^{(i)},a^{(i)},\beta). \]
\noindent
The model might be a logistic regression $\Pm(c=1|x_1,a,\beta)=\sigma( (x_1 \otimes a)^T \beta)$.
From a confounding point of view, everything is fine,  it doesn't matter that  $x_2$ is ignored. 
In practice this means that if the estimate $\hat{\beta}_1$ is good it means that the `true' CTR for any given $x_1$ and action $a$ is indeed given by $\Pm(c=1|x_1,a,\hat{\beta}_1)$.  Similarly, the model can be used to find the optimal $a$ for any given $x_1$ which is achieved using an epsilon greedy policy starting on Day 1.

\noindent
\paragraph{Day 1:} the reco team then deploys on Day 1 the following policy:

\[
\pi_1(a|x_1) = (1-\epsilon
) \boldsymbol{1}\{a = {\rm argmax}_{a'} \Pm(c=1|x_1,a',\hat{\beta}_0) \}+ \frac{\epsilon}{A
}
\]
\sloppy

\noindent
where $\epsilon$ controls an epsilon greedy policy.  This new policy  produces an uplift and everyone is happy.  The reco team also collect some new data for Day 1 $\mathcal{D}_1=\{c^{(i)}, x_1^{(i)}, x_2^{(i)} , a^{(i)}\}_{i=L_0+1}^{L_1}$.  
The reco team then decides that maybe they could also incorporate the features $x_2$ into the model, so at the end of Day 1 they then fit:

\[
\hat{\beta}_1 = {\rm argmax}_\beta \sum_{i=L_0+1}^{L_1} \log \Pm(c^{(i)}|x_1^{(i)},x_2^{(i)}, a^{(i)},\beta).
\]
\noindent
The model again might be a logistic regression $\Pm(c=1|x_1,a,\beta)=\sigma( (x_1 \otimes x_2 \otimes a)^T \beta)$.  There is no confounding, and indeed there is better personalization because now $x_2$ is used.  
Again, this means that the `true' CTR for a user arriving with features $x_1,x_2$ and action $a$ is  estimated by 
$\Pm(c|x_1,x_2,a,\hat{\beta}_1)$.

\noindent
\paragraph{Day 2} the reco team then deploy the following policy:

\[
\pi_2(a|x_1,x_2) = (1-\epsilon
) \boldsymbol{1}\{a = {\rm argmax}_{a'} \Pm(c=1|x_1,x_2,a',\hat{\beta}_1) \} + \frac{\epsilon}{A
}.
\]

\noindent
The new policy is more personalized (as it now uses both $x_1$ and $x_2$) and  produces an uplift, but for technical reasons it is decided that incorporating the extra complexity of using $x_2$ is not worth the extra engineering cost.
So, the reco team decide that in the future the model will return to  only use $x_1$ i.e. they re-apply the methodology they applied at the end of Day 0.  It seems like the same methodology should work again, but will it?  The data collected is $\mathcal{D}_2=\{c^{(i)}, x_1^{(i)}, x_2^{(i)}, a^{(i)} \}_{i=L_1+1}^{L_2}$.    They use this data to estimate:

\[
\hat{\beta}_2 = {\rm argmax}_\beta \sum_{i=L_1+1}^{L_2} \log \Pm(c^{(i)}|x_1^{(i)}, a^{(i)},\beta) 
\]
\noindent
Unfortunately, the model $\Pm(c=1|x_1,a,\hat{\beta}_2)$ is now confounded, this means that the true CTR for a given $x_1$ and action $a$ is \emph{not} given by $\Pm(c=1|x_1,a,\hat{\beta_2})$, similarly selecting the $a$ that maximizes the click probability will  \emph{not} give the best policy.

\noindent
\paragraph{Day 3:} the reco team deploy:
\[
\pi_3(a|x_1) = (1-\epsilon
) \boldsymbol{1}\{a = {\rm argmax}_{a'} \Pm(c=1|x_1,a',\hat{\beta}_2) \} + \frac{\epsilon}{A
}
\]

\begin{figure}[h]
    \centering
    \begin{subfigure}[b]{0.39\textwidth}
        \centering
        \includegraphics[width=\textwidth]{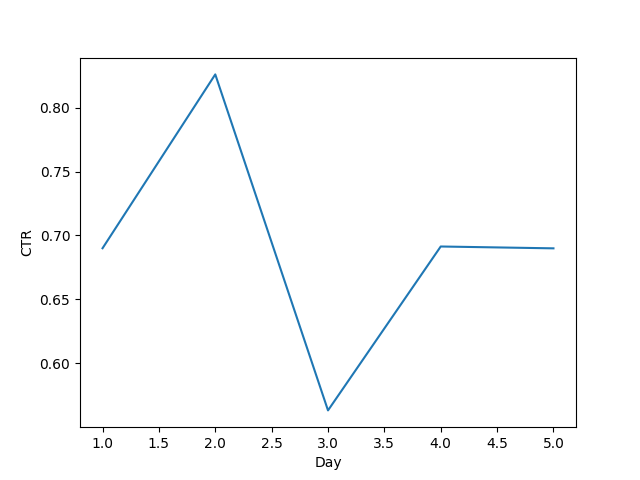}
        \caption{The standard practice of feature engineering causes confounding on Day 3.  Exactly the same training procedure is used on Day 1, 4 and 5 as Day 3, but on the log for Day 3 (Day 2) the policy implements $x_2 \rightarrow a$ causing $x_2$ to not be ignorable.}
        \label{fig:a}
    \end{subfigure}
    \hfill 
    \begin{subfigure}[b]{0.39\textwidth}
        \centering
        \includegraphics[width=\textwidth]{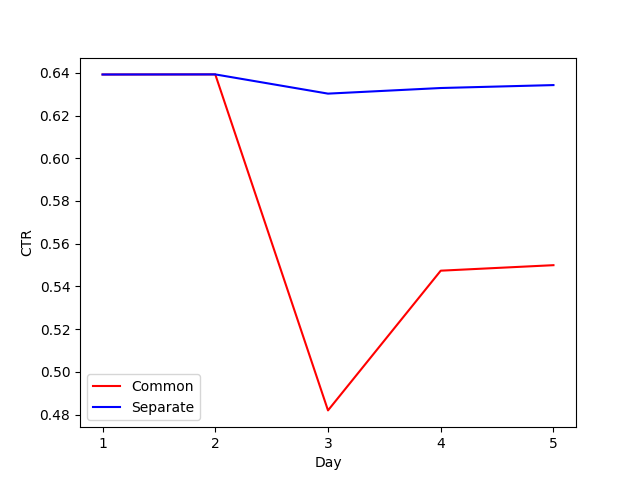}
        \caption{Demonstrates how A/B testing with common training can entrench confounding effects
        The A/B test starts on Day 1, where A uses only $x_1$ and B uses $x_1,x_2$.  Both A and B are trained on a combined log resulting in A suffering from confounding that persists as long as the A/B test runs.  The plot shows two approaches to A/B testing with separate training data (blue) and combined (red), only population A suffers confounding and only population A is shown.
        }
        \label{fig:b}
    \end{subfigure}
    \label{fig:featureeng}
\caption{Confounding simulations showing the effect of feature engineering and A/B testing}    
\end{figure}

\noindent
The click model is confounded and consequently the wrong preferred action is selected for some contexts resulting in a lower average click through rate.  On subsequent days, (Day 4, Day 5), the confounding disappears again, because the model is only trained on the previous day and from Day 3 onwards, the recommendations are determined only by $x_1$.  Figure \ref{fig:a} shows a simulated demonstration of this behavior where confounding impacts the Day 3 CTR but it returns on Day 4 and 5.  This suggests that one solution to mitigate confounding is to simply wait, unfortunately this solution is not generally applicable.  Consider the same scenario but now the feature engineering performance is being measured using A/B testing and both A and B are trained on a common log as shown by the red line in Figure \ref{fig:b}, the  practice of  having each policy training on its own log as shown by the blue line  remedies the confounding problem but reduces the sample size. 

The experiments in Figure \ref{fig:a} and Figure \ref{fig:b} have a similar setup. 
Both $x_1$ and $x_2$ are categorical variables with 5 states and $a$ is a categorical with 10 states.  The simulations use 400 000 samples.  In Figure \ref{fig:b} the A/B test starts on Day 2, causing confounding starting on Day 3.  Population A is the model that only has access to $x_1$, only population A suffers from confounding as such only the CTR of population A is shown.  Code for both experiments is available \href{https://colab.research.google.com/drive/1hvt_skstFsLdsShEaWLsKlWXN0ZF_lQt?usp=sharing}{\color{blue}{\underline{here}}}.

\section{Confounding due to Feature Engineering and Modularization}
\begin{figure}
    \centering
\begin{tikzpicture}[
    node/.style={
        circle,
        draw,
        minimum size=1cm,
        font=\sffamily\small
    },
    arrow/.style={
        -Stealth,
        thick
    }
]

\node[node] (xp) at (0, 2) {$x'$};
\node[node] (xpp) at (2, 2) {$x''$};
\node[node] (a) at (-2, 0) {$a$};
\node[node] (c) at (0, -2) {$c$};
\node[node] (s) at (2, -2) {$s$};

\draw[arrow] (xp) -- (c);
\draw[arrow] (xp) -- (s);
\draw[arrow] (xp) -- (xpp);
\draw[arrow] (xpp) -- (c);
\draw[arrow] (xpp) -- (s);
\draw[arrow] (a) to[out=0, in=180] (c); 
\draw[arrow] (a) to[out=0, in=180] (s); 
\draw[arrow] (xp) -- (a);
\draw[arrow] (xpp) -- (a);
\end{tikzpicture}

    \caption{Causal graph for the post click sale models.  If the sale model only is able to see $x'$ and the click model is only able to see $x''$ then the two models confound each other.}
    \label{fig:clicksale}
\end{figure}
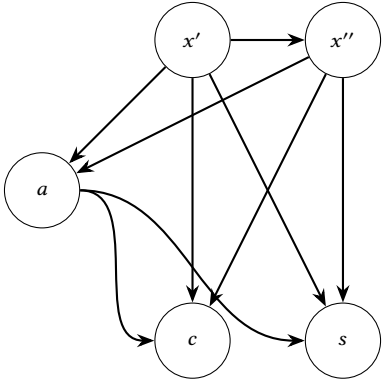
Modularization is an important engineering practice, but when combined with feature engineering it can also lead to confounding.  Consider the situation where there is a separate sale and click model.  Let $c$ be a click, $s$ be a sale, $a$ be an action and $x$ be the context, the goal is to maximize post click sales.  This can be achieved by solving

\begin{align*}
a^* & = {\rm argmax}_a P(c=1,s=1|a,x) \\& = {\rm argmax}_a P(s=1|a,x,c=1)P(c=1|a,x)    
\end{align*}

\noindent
Now, consider that a different team build the click and the sale model and they both do feature engineering and produce different feature sets $x'$, and $x''$, so instead the delivered action is:

\[
a^* = {\rm argmax}_a  P(s=1|a,x',c=1)P(c=1|a,x'')
\]

\noindent
This procedure looks sound from the point of view of the individual models, but when the individual models are estimated on real data the above procedure produces sub-optimal actions due to each model being confounded.  The causal graph is shown in Figure \ref{fig:clicksale}, training the click model and sales model individually results in either $x'$ or $x''$ being an unobserved (or more accurately ignored) confounder.  This highlights a dilemma.  As the post click sales model is trained on a log where clicks have already occurred it is a much smaller dataset and hence there is a risk of higher variance in the estimation.  Conventional machine learning wisdom suggests that $x'$ should be simpler than $x''$ as a way to reduce this variance in the sales model, but having $x'$ and $x''$ different leads to confounding.  

A similar situation can occur when there are two decisions, let's say $d$ is the decision to display a recommendation and $a$ is the recommendation and our problem is to maximize clicks $c$, then a correct solution (neglecting exploration) is:

\begin{align}
\hat{\beta} = {\rm argmax}_\beta \sum_{i=1}^N \log P(c_i|a_i,d_i,x_i,\beta)
\label{bidreco}    
\end{align}

\[
\pi(a,d|x) =  \boldsymbol{1} \{a,d = {\rm argmax}_{a',d'}  P(c=1|a',d',x,\hat{\beta}) \} 
\]

\noindent
However perhaps  the decision of $a$ and $d$ must be made separately and without knowledge of the other part of the system.  A common practice is to fit two models, $P(c|a,x')$ and $P(c|d,x'')$, and let $\pi(a,d|x) = \pi(a|x') \pi(d|x'')$, where $\pi(a|x')=\boldsymbol{1} \{ a={\rm argmax}_{a'} P(c=1|a,x')\}$, and $\pi(d|x')=\boldsymbol{1} \{ a={\rm argmax}_{d'} P(c=1|d,x'')\}$.  Similar to the above example each of these models confound each other.  

Avoiding confounding requires fitting the model in Equation 3, and then if engineering limits require the policy to be simplified to $\pi(a,d|x)=\pi_\Xi(a|x')\pi_\Gamma(d|x'')$, where $\Xi$ and $\Gamma$ parameterize the implementable policies.  Policy learning can be used to find:

\[
{\rm argmax}_{\Xi,\Gamma} E_{a \sim \pi_\Xi(a|x') d \sim \pi_\Gamma(d|x'') x,x',x'' \sim P(x,x',x'') }P(c=1|a,d,x,\hat{\beta}).
\]

\noindent
Algorithms such as REINFORCE \cite{williams1992simple} can be used to target this type of objective.


\section{Past Work on Confounding for Recommender Systems}
Recommender systems by construction have access to all covariates that lead to past recommendations, hence confounding can in principle be avoided simply by adjusting for all non-ignorable covariates.  

The study in \cite{xu2023instrumental} develops personalization algorithms for cases where confounders are truly unobserved using an instrumental variable approach.  As past recommendations are always a function of observed information this does not apply to recommendation.  Computational advertising can handle market conditions (price of inventory) by making the action the bid, or by using intention to treat, which reformulates the problem without any unobserved confounders.

The study in \cite{jadidinejad2021simpson} does not use click models to build a recommender system but rather applies to standard collaborative filtering datasets.  It is rather difficult to map this work to the practices in a real world recommender system.

The study in \cite{jeunen2023offline} considers estimating the reward of a policy using inverse propensity score based estimators when both the propensity and covariates that determine the propensity are missing.  

Propensity score methods are relevant to confounding, there are two broad approaches.  Balancing scores \cite{rosenbaum1983central} use the balancing score as a (usually) low dimensional substitute for a covariate in order to estimate an unconfounded model.  In contrast, inverse propensity score estimators (IPS) avoid using a model and instead estimate the expected utility of a new policy \cite{joachims2018deep}.  Variants of this estimator typically make trade-offs in terms of bias and variance of the expected utility, common variants include clipping, self-normalized importance sampling and doubly robust.
Both balancing scores for models propensity scores for policy estimators avoid problems of confounding.  The trade-offs in using estimators directly of the policy utility, rather than a likelihood based approach for estimating a model are rather complicated (see \cite{sims2006example}) and beyond scope.  Broadly, both methods do something that might be considered unnecessary; model based approaches must estimate the reward of every action, IPS based methods must estiamte the expected utility (which is not needed to select an action).  IPS based estimators can have very high variance and violate the likeihood principle \cite{berger1988likelihood}, but they are much more practical in multi-turn problems \cite{sakhi2025practical}.

\bibliographystyle{ACM-Reference-Format}
\bibliography{Rohde_refs}

\appendix




\section{Are These Practices Really `Pervasive'?}



Real recommender systems are built from many sub-models.  Although the details of real systems are not publicly known, it is not a secret that many systems use at least two stages \cite{borisyuk2016casmos,yi2019sampling} and separate the reward into components such as clicks and sales \cite{vasile2017cost}.  To avoid confounding requires a lot of discipline, either using the same features in all sub-models or another more sophisticated approach.  Given that no paper describes this practice, it is reasonable speculation that most tech companies do not implement a strategy to avoid confounding.  Moreover, there are real costs in making avoiding confounding a priority.  Working legacy systems will need significant modification.  Parameter estimation accuracy may suffer in some sub-models (that have their feature dimension increased).  Also note that many tech companies prioritize building highly personalized recommendation engines using many features perhaps based on deep learning models. Often the pursuit of this goal will be at the expense of making sure that all sub-models are absolutely free from confounding.

Similarly, the temporal confounding effects of adding or removing a feature into a sub-model are not documented in the literature.  While surely some practitioners are aware of these concerns, the lack of a systematic discussion, strongly suggests that it is neglected by most tech companies.

It is hard to know the A/B testing practices used at different tech companies, and again, there is no systematic treatment of the concept applied to recommendation in the literature (where models are re-trained).  Our reasonably informed speculation is that some A/B tests involve models training on a common log producing another source of confounding.




\section{Strategies for Avoiding Confounding}

Some strategies for avoiding confounding remain research questions, but the following ideas should be implemented where possible:

\begin{itemize}
    \item A/B Testing should involve each candidate recommender system training exclusively on its own log.
    \item A feature can be added to (all models) within a system without any concern about confounding.
    \item The removal of a feature from (all models) within a system will cause confounding.  Simply, waiting for the confounded log to move outside the training window is one possible approach.  Another approach is to use the backdoor rule, but this is likely very unappealing to practitioners.
    \item Adding a feature into a sub-model will improve the performance of that sub-model, but confound all other models.  This should give pause to thought to improving only one sub-model.
    \item Putting the same features in all sub-models might well be unacceptable from a model fitting point of view (some models may estimate poorly if the feature dimension becomes larger).  One  approach is to use the 
balancing scores (see  \cite{rosenbaum1983central}) which is a score $b(x_2)$ such that $x_2 \rightarrow a$ can be replaced with $b(x_2) \rightarrow a$, using $b()$ as a feature can help modularize, reduce variance and avoid confounding.  However, the first step is for practitioners to simply recognize that confounding is a potentially serious problem in real recommender systems.

\end{itemize}

\end{document}
\endinput